\documentclass[letterpaper]{article} 
\usepackage{amssymb}
\usepackage{etoolbox}
\usepackage{sequential}  
\usepackage{times}  
\usepackage{helvet}  
\usepackage{courier}  
\usepackage[hyphens]{url}  
\usepackage{graphicx} 
\usepackage{natbib}  

\usepackage{caption} 
\usepackage{algorithm}
\usepackage{algpseudocode}
\usepackage{amsmath}

\newcommand {\pop} {N}
\newcommand {\feat} {x}
\newcommand {\Pfeat} {\mathbf{x}}
\newcommand {\featSpace} {\mathcal{X}}
\newcommand {\pol} {\pi}
\newcommand {\Pact} {\mathbf{a}}
\newcommand {\act} {a}

\newcommand {\Pout} {\mathbf{y}}
\newcommand {\out} {y}

\usepackage{newfloat}
\usepackage{listings}
\DeclareCaptionStyle{ruled}{labelfont=normalfont,labelsep=colon,strut=off} 
\floatstyle{ruled}
\newfloat{listing}{tb}{lst}{}
\floatname{listing}{Listing}
\title{Sequential Cohort Selection under Uncertainty}
\author {
    Hortence Yiepnou,
    Christos Dimitrakakis
}
\affiliations {
  University of Neuch\^atel\\
    \{hortence.yiepnou, christos.dimitrakakis\}@unine.ch
}

\usepackage{bibentry}

\begin{document}

\maketitle

\begin{abstract}
We study the problem of fair cohort selection under uncertainty, motivated by university admissions where applicant outcomes  are only partially observed. We consider both a one-shot setting, where a fixed policy is applied to a population, and a sequential setting, where policies are updated over time using data from previous admission years.
We propose a policy optimization framework that combines probabilistic modeling of outcomes with policy gradient methods, supporting both logistic and neural network policies. In the sequential setting, the approach jointly updates the policy and the underlying models to adapt to evolving applicant populations.
Experiments on a simulator grounded in real admission data show that adaptive policies substantially outperform static baselines in term of expected utility, especially under higher admission costs. Neural policies consistently achieve higher utility and adapt more effectively than simpler models, while maintaining favorable fairness properties over time. Our results demonstrate the importance of adaptivity and model expressiveness for decision-making under uncertainty.
\end{abstract}

\section{Introduction}
Decision-making under uncertainty is central to many high-stakes applications, including university admissions, hiring, or resource allocation \cite{gibney2007decision, schmitt2007selection, yoshimura2006decision,manski2017improving}. In such settings, decision makers must select individuals based on imperfect information about their future outcomes, while also considering fairness and resource constraints. This creates a fundamental tension between utility maximization, uncertainty, and equity considerations \cite{dwork2012fairness, hardt2016equality}.

University admissions provide a particularly compelling example. Admission committees aim to select applicants who are most likely to succeed academically, yet outcomes such as future grades or performance are uncertain at decision time. At the same time, institutions face capacity constraints and are increasingly required to satisfy fairness criteria, such as demographic parity (DP) or equality of opportunity (EO) \cite{zafar2017fairness, chaibub2020causal}. These challenges are further compounded by the fact that the applicant pool evolves over time and only the results of admitted students are observed, introducing selection bias in training data \cite{heckman1979sample, mehrotra2021mitigating}. Importantly, this limitation is intrinsic to real-world decision-making settings: The outcomes of rejected applicants are not observed and therefore cannot be directly used for learning. Rather than requiring such counterfactual information, our approach operates under this partial observability by modeling uncertainty in outcomes through Bayesian inference and approximating the applicant distribution using a generative model such as the Conditional Tabular Generative Adversarial Network (CTGAN) \cite{xu2019modeling}. While this does not explicitly recover missing outcomes for rejected individuals, it enables the policy to be optimized under uncertainty using the available data. 

Most existing approaches address these challenges in isolation. Traditional methods rely on static ranking or threshold-based rules, which do not explicitly model uncertainty, or focus on fairness-aware prediction, without accounting for downstream decision-making and resource constraints \cite{garnett2010bayesian, kleinberg2018algorithmic}. More work on policy learning often assumes a fixed data distribution and does not consider the dynamic nature of real-world decision processes, where both the population and the model evolve over time \cite{ athey2021policy, kleine2022meritocracy,gupta1958selecting}.

In this work, we propose a framework for learning decision policies under uncertainty in dynamic populations. Our approach models both the feature distribution of applicants and the outcome distribution conditional on decisions, allowing the decision maker to reason about uncertainty in a principled way. We formulate the selection problem as the optimization of an expected utility function, which captures both predicted outcomes and admission costs, and measures fairness constraints such as equitable representation across demographic groups. Importantly, the utility is non-additive across individuals, reflecting the coupling induced by shared resource constraints and the dependence of model updates on the selected cohort.

To optimize the policy, we adopt a policy gradient approach, in which the admission decision is modeled probabilistically and learned directly from the data \cite{wu2023predictive}. While a simple logistic policy can provide a strong baseline, we also consider a more expressive neural network policy, which can capture nonlinear relationships between applicant features and optimal decisions. We study this framework in two complementary settings. In a single-stage (one-shot) setting, the policy is learned on a fixed population and evaluated on new samples from the same distribution. In a multi-stage setting, we model a more realistic scenario where the applicant distribution is unknown and evolves over time. We use a generative model to simulate incoming applicants and iteratively update the policy, the applicant distribution model, and the outcome model as new data become available.

\paragraph{Contribution.}
This paper makes the following contributions.
\begin{itemize}
    \item We propose a novel policy-gradient-based framework for cohort selection from unknown populations, enabling the optimization of complex, non-linear selection policies.
    \item We extend the framework to a multi-stage setting where the applicant population evolves over time. Using a generative model, we simulate new applicants and iteratively update both the policy and the outcome model. We personalize selection by assigning individual-specific probabilities under selection framework and also measure fairness constraints such as DP and EO.
    \item We demonstrate the effectiveness of our approach on realistic data by comparing it with standard baselines. We analyze the impact of key parameters and show how learned policies balance utility and fairness across different settings.
\end{itemize}

\section{Related Work}

Our work is at the intersection of decision-making under uncertainty, fairness in machine learning, and policy learning in dynamic environments. We briefly review the most relevant strands of literature.

\textbf{Decision-making under uncertainty.} Practical selection processes are inherently defined by decision-making under uncertainty, as institutional policies are often defined and made transparent before the specific characteristics of the applicant pool are observed~\cite{kochenderfer2015decision,johnson2010decision,trimmer2011decision,schultz2010decision,soroudi2013decision,faucheux1995decision,preuschoff2013decision}. This requires a shift from deterministic selection \cite{manski2004measuring} to a probabilistic framework in which the candidate pool is modeled as a realization of a potentially unknown probability distribution $P$. In such environments, the decision-maker must optimize an expected utility objective that accounts for both the stochastic nature of applicant arrivals and the predictive uncertainty regarding their future outcomes \cite{johnson2010decision,gardenfors1982unreliable}. Our framework addresses this by formalizing the selection task as an optimization problem on a  policy $\pi$, where the goal is to maximize the expected utility $U(\pi, P)$ across all possible population realizations. 

\textbf{Fairness in Selection.} Fairness in automated decision-making has been a growing area of interest, with multiple formalizations proposed to ensure equitable outcomes across different demographic groups. It is addressed by~\cite{zhang2021fair,zhang2015fairness}, particularly in settings where class labels are censored. The authors propose a new fairness framework that explicitly accounts for censored data, ensuring that decisions do not disproportionately disadvantage certain groups. For instance, DP mentioned by \cite{caton2024fairness,chouldechova2017fair,corbett2017algorithmic,dwork2012fairness,dwork2018group,dwork2018individual} ensures that the selection probabilities are proportional between groups, regardless of the underlying performance differences. However, DP has been criticized for potentially disregarding individual qualifications. To address this, \emph{the EO} introduced by~\cite{hardt2016equality} ensures that individuals with similar qualifications receive comparable treatment. Our work incorporates both fairness notions as evaluation metrics for policies optimized purely for utility, allowing us to observe empirically whether utility-maximizing policies incur fairness costs, in contrast to ~\cite{bagriacik2024multiple} who optimize a classification fairness metric directly. 

The tension between fairness and utility has been a major concern in algorithmic decision making ~\cite{corbett2017algorithmic}. Their studies have shown that enforcing strict fairness constraints can lead to reduced overall utility, raising questions about the optimal balance between fairness and efficiency \cite{berk2021fairness}. Our study explores a weighted combination of fairness penalties, allowing decision makers to well balance fairness and utility according to policy objectives. Additionally, we investigate the use of a maximum operator formulation to ensure that the most severe fairness violations are mitigated.

Classical meritocratic models often assume that individual merit is static and observable \cite{mas1995chapter}. This assumption has been challenged on normative grounds, most notably by Rawls \cite{rawls1971theory}, who treats natural talents and abilities as morally arbitrary and therefore not a legitimate basis for desert. However, the Expected Marginal Contribution (EMC) framework  introduced a way to quantify merit by assessing an individual's specific contribution to the aggregate utility of a subset \cite{kleine2022meritocracy}. While prior implementations of EMC were limited to single-shot settings , we integrate EMC into a policy framework that accounts for multi-shot setting and population uncertainty, thereby, preserving meritocratic principles even when the applicant pool is not fully observed.

\textbf{Policy learning and dynamic population.}
Our approach is closely related to policy learning, where the goal is to learn a mapping from features to actions that maximizes expected outcomes \cite{ athey2021policy}. In particular, we adopt a policy gradient method, a standard technique in offline reinforcement learning to optimize stochastic policies \cite{levine2020offline,sutton2018reinforcement}. Although policy gradient methods have been used in sequential decision-making problems, many existing works assume a fixed environment or fully observed rewards \cite{parisi2014policy,wang2016knowledge}. In contrast, our setting involves partial observability, since outcomes are only observed for selected individuals. Additionally, we consider both linear (logistic) and nonlinear (neural network) policy parameterizations to study the impact of model expressiveness on decision quality.

At the same time, many real-world decision problems involve non-stationary or evolving populations, where the data distribution changes over time. Recent work has explored the use of generative models, such as GANs, to simulate and adapt to changing data distributions \cite{goodfellow2014generative}. Our framework combines these ideas by using a generative model to simulate applicant populations and iteratively updating both the outcome model and the decision policy in multiple stages, thereby, capturing the dynamic nature of real-world decision processes, which is often overlooked in static settings.

\section{Setting and Notation} 

In this section, we formally define the problem, introduce the notation, and specify the models used throughout the paper.

\textbf{Population and features.} We consider a selection process where a decision-maker must select a cohort from a population of $\pop$ individuals. Each individual $i$ is described by the features $\feat_i \in \featSpace$. For the running example of college admissions, these  can include socio-demographic attributes ( age, gender, citizenship, etc.) as well as performance metrics (high school GPA, science points, etc.). 
We denote the complete population's features as $\Pfeat \in \featSpace^\pop$. While the candidate population is unknown, we assume that we have access to a probability distribution $P$ over possible populations. This can be achieved through probabilistic inference given past admission data.

\textbf{Actions and outcomes.} We define the acceptance policy $\pol$ before we see the population. This reflects practical scenarios where policies must be implemented without the knowledge of the population. The form of admission policy $\pol$ that we will focus on generates a probability of accepting each individual independently, based on their features. That is, for every individual $i$, it calculates a conditional probability $\pol(a_i |  x_i)$, where $a_i = 1$ means that the individual $i$ is accepted and $a_i = 0$ means that they are rejected. For a given population feature matrix $\Pfeat$, the policy generates an acceptance vector $\Pact$ so that:
\[
\pol(\Pact \mid \Pfeat) = \prod_{i=1}^\pop \pol(\act_i \mid \feat_i).
\]
This policy then defines how the selection will be made for \emph{any} possible observed population $\Pfeat$.

Once a person $i$ is accepted, we can observe the outcome $\out_i$ of our decision: in the university admissions case, this can be the grades achieved in different subjects taken, with $\out_{i,j}$ being individual $i$'s outcome in the $j$-th course, with the course results of  population $\Pfeat$ denoted by $\Pout$.

\textbf{Utility under uncertainty.} Our policy optimization target is to maximize a utility function in expectation under a distribution over possible applicant populations.
\footnote{Even if we do not have access to the true population distribution, in this paper we use probabilistic inference to estimate it.}  We can then update our distribution using data from each year's accepted students. The distribution thus, has two components. First, $P(\Pfeat)$: the probability of having a specific candidate population. Secondly $P(\out_i \mid x_i, a_i = 1)$, the probability of observing a specific outcome for an accepted individual.\footnote{The outcomes for rejected individuals remain unobserved.} The utility function $u(\Pact, \Pout)$ is defined in terms of the selection \(\Pact\) and the outcomes \(\Pout\) obtained. 
For concreteness, we define the following log-linear utility function:
\begin{equation}
\label{eq:U_ay}
u(\Pact, \Pout) = \sum_{k=1}^K \log\left(\sum_{i \in N} a_i \cdot y_{i,k}\right) - c \cdot \|\Pact\|_1,
\end{equation}
where $y_{i,k} $ is the outcome of candidate $ i $ in course $ k $, $ \|\Pact\|_1 $ is the total number of candidates selected and $ c $ is the cost of selecting a candidate. The logarithmic term enforces diminishing returns on individual performance, encouraging a balanced and high-performance cohort, and the second term $- c \cdot \|\Pact\|_1$
 penalizes selecting too many candidates. Instead of enforcing a fixed admission quota, we incorporate a soft budget constraint. Using this formulation, the decision-maker can easily  balance the trade-off between admitting more candidates and maintaining high expected outcomes, without requiring a predefined admission rate. In practice, strict quota may be difficult to specify in advance, especially under uncertainty about applicant quality or future outcomes. 

 Although we want our selection to be optimal in expectation, we have to consider the uncertainty in the features and outcomes of the population. 
When outcomes \( \Pout \) are unknown, but the decision maker observes features \( \Pfeat \), the utility of taking an action \( \Pact \) is the expected utility (reflecting the uncertainty about \( \Pout \)) over possible outcomes:
\begin{equation}
\label{eq:U_ax}
U(\Pact, \Pfeat) \triangleq \mathbb{E}_{\Pout \sim P(\Pout \mid \Pfeat)} \left[ u(\Pact, \Pout) \mid \Pfeat , \Pact\right].
\end{equation}
As the decision maker does not observe the population $\Pfeat$, we have to optimize the policy in terms of the expected utility under a distribution $P$ of possible populations; the expected utility of policy $\pol$ is given by:
\begin{equation}
\label{eq:U_pi}
U(\pol, P) \triangleq \mathbb{E}_{\Pfeat \sim P(\Pfeat)} \mathbb{E}_{\Pact \sim \pol(\Pact \mid \Pfeat)} \left[ U(\Pact, \Pfeat) \right].
\end{equation}
 The primary goal of the decision maker is to maximize the expected utility ($\max_{\pol} U(\pol, P)$). This objective accounts for three layers of uncertainty: the composition of the applicant pool, the selection decisions of the policy, and the academic outcomes obtained.
While utility maximization is not necessarily aligned with fairness considerations, \cite{kleine2022meritocracy} showed that it can satisfy a type of meritocratic fairness related to stability. 

\section{Policy Optimization}
\label{sec:algorithm}
Given a population distribution $P(\Pfeat)$ and an outcome distribution $P(\Pout | \Pact, \Pfeat)$, we must find an optimal policy. We do this using policies with some parameters $\theta$ through gradient ascent or descent (which will be specified later in this section), for any arbitrary utility function $u(\Pact, \Pout)$. 

\textbf{Policy Structures and Parameters.}
The simplest policy structure we have, denoted by \( \pol_1(\Pact \mid \Pfeat) \), uses a linear-sigmoid function of the features:
\begin{equation}
\pol_1(\Pact \mid \Pfeat) = \prod_{i=1}^\pop \pol_1(a_i \mid x_i),
\qquad
\pol_1(a_i = 1 \mid x_i) = \frac{e^{(\theta_1^\top x_i)}}{1 + e^{(\theta_1^\top x_i)}},
\end{equation}
where \( x_i \) is the feature vector for the candidate \( i \), and \( \theta_1 \) is the parameter vector that defines this policy.  

The second policy structure, $\pol_2(\Pact \mid \Pfeat)$, uses a neural network that outputs admission probabilities for each candidate conditioned on the features.
\begin{equation}
\pol_2(\Pact \mid \Pfeat) = \prod_{i=1}^\pop \pol_2(a_i \mid x_i), \qquad \pol_2(a_i = 1 \mid x_i) = f_{\theta_2}(x_i),
\end{equation}
where $f_{\theta_2}(\cdot)$ is a neural network parametrized by $\theta_2$ with sigmoid output activation to model admission probabilities. Unlike the first policy, this architecture allow to model complex nonlinear relationships between features and admission decisions. The architectures of this second policy will be described in section \ref{sec:exp}.

In the remainder, we will use \( \pol(\Pact \mid \Pfeat) \) and  \(\theta \)  when there is no need to distinguish between the two policy types.

\textbf{Gradient Calculation.} This optimization problem is not a typical supervised learning problem. Consequently, we must explicitly approximate the expected utility gradient. We start with the expected utility of the policy.
\begin{equation}
U(\pol, P) = \sum_\Pfeat \sum_\Pact P(\Pfeat) \pol(\Pact \mid \Pfeat) U(\Pact, \Pfeat),
\end{equation}
where $U(\Pact, \Pfeat)$, given by equation \ref{eq:U_ax}, is approximated by sampling from the outcome prediction model  $P(\Pout \mid \Pact, \Pfeat)$.

To optimize the policy using gradient ascent, we need to compute the gradient of \( U(\pol, P) \) with respect to the parameters \( \theta \) of the policy. This gradient provides a direction for adjusting \( \theta \) to maximize \( U(\pol_\theta, P) \):

\begin{align}
 & \frac{\partial U(\pol_\theta, P)}{\partial \theta_i} = \sum_\Pfeat P(\Pfeat) \sum_\Pact \frac{\partial }{\partial \theta_i}  U(\Pact, \Pfeat) \pol_{\theta}(\Pact \mid \Pfeat) \notag \\
&= \sum_\Pfeat P(\Pfeat) \sum_\Pact \pol_{\theta}(\Pact \mid \Pfeat) U(\Pact, \Pfeat) \sum_{j=1}^N \frac{\partial \log(\pol_{\theta}(a_j\mid x_j))}{\partial \theta_i}  \\
&=  \sum_\Pfeat P(\Pfeat) \sum_\Pact  \pol_{\theta_1}(\Pact \mid \Pfeat) U(\Pact, \Pfeat) \bigg[ \sum_{j=1}^N x_{i,j} \notag\\
 & \quad  [\mathbf{1}_{a_j=1}  \pol_{\theta_1}(a_j =0 \mid x_j) \notag  -  \mathbf{1}_{a_j=0}  \pol_{\theta_1}(a_j =1 \mid x_j) ]\bigg].\notag
\end{align}


In practice, summing over all possible \( \Pfeat \) and \( \Pact \) is intractable, we approximate these expectations using Monte Carlo sampling. We first sample \( n_\Pfeat \) instances \( \Pfeat^{(k)} \sim P(\Pfeat) \) from the population distribution. For each sampled \( \Pfeat^{(k)} \), we then sample \( n_\Pact \) actions \( \Pact^{(k,m)} \sim \pol_{\theta}(\Pact | \Pfeat^{(k)}) \) from the policy. Using these samples, the gradient is approximated as:


\begin{align}
 \frac{\partial U(\pol_\theta, P)}{\partial \theta_i} & \approx \frac{1}{n_\Pfeat} \sum_{k=1}^{n_\Pfeat} \frac{1}{n_\Pact} \bigg[\sum_{m=1}^{n_\Pact} U(\Pact^{(k,m)}, \Pfeat^{(k)}) \notag \\
 & \sum_{j=1}^N \frac{\partial }{\partial \theta_i} \log \pol_\theta(a_j^{(k,m)} \mid x_j^{(k)})\bigg],
\end{align}

or more specifically for the first policy,
\begin{align}
 \frac{\partial U(\pol_{\theta_1}, P)}{\partial \theta_{i_1}} &\approx \frac{1}{n_\Pfeat} \sum_{k=1}^{n_\Pfeat} \frac{1}{n_\Pact} \sum_{m=1}^{n_\Pact} U(\Pact^{(k,m)}, \Pfeat^{(k)}) \notag\\& \sum_{j=1}^N x_{i,j}^{(k)} \bigg[ \mathbf{1}_{a_j^{(k,m)}=1}  \pol_{\theta_1}(a_j^{(k,m)} =0|x_j^{(k)}) \notag  \notag\\ &-  \mathbf{1}_{a_j^{(k,m)}=0}  \pol_{\theta_1}(a_j^{(k,m)} =1|x_j^{(k)}) \bigg].
\end{align}

$n_\Pfeat$ is the number of population samples, \( n_\Pact \) is the number of action samples per \( \Pfeat \), and
\( \pol_{\theta_1}(a_j^{(k,m)} =1|x_j^{(k)}) = \frac{1}{1 + \exp(\theta_1^T \cdot x_j^{(k)})}.\)

As mentioned previously, the objective is to maximize the expected utility and this is done by iteratively updating the parameter $\theta$ using gradient ascent:
\begin{equation} 
\theta^{t+1} \leftarrow \theta^t + \eta \cdot \nabla_\theta U(\pol_\theta, \Pfeat).
\end{equation}
where $\eta$ is the learning rate

 When using neural networks, this optimization is performed via stochastic gradient descent (SGD) or one of its variants (e.g., Adam), using automatic differentiation frameworks such as PyTorch or TensorFlow. To match the framework's default of minimizing losses, we define the loss function as the negative expected utility:
 $$
 \mathcal{L}(\theta) = - \mathbb{E}_{\pol_\theta(\Pact \mid \Pfeat)}[U(\Pact, \Pfeat ) \cdot \log \pol_\theta(\Pact \mid \Pfeat)],
 $$
 and minimize it using backpropagation:
 $$
 \theta^{t+1} \leftarrow \theta^t - \eta \cdot  \nabla_\theta \mathcal{L}(\theta),
 $$
 which is mathematically equivalent to gradient ascent on the utility objective.
 
\textbf{Batch and Sample Sizes.}
In our setting, the utility \emph{cannot} be broken down into a sum over individuals, as it is a utility over sets. This means that the \emph{batch size} is equivalent to the size of the applicant pool, i.e. the number of individuals in each sampled $\Pfeat^{(k)}$. In addition,  the number of samples for the population (\( n_\Pfeat \)) and actions (\( n_\Pact \)) may not be the same. Their choice influences the variance of the estimate.

\textbf{Algorithms.} The practical implementation of the proposed optimization procedures for both one-shot and multi-stage settings is summarized in Appendix \ref{app:algo}. These algorithms detail the Monte Carlo estimation of the policy gradient, the sampling of actions, and the iterative update of the policy parameters. While the core ideas follow directly from the formulations presented above, the appendix provides a structured view of the full training procedure for clarity and reproducibility. Additional implementation details are provided in Sections~\ref{sec:exp} and~\ref{mm}, which describe the experimental setup and the sequential learning framework in practice.

\section{Fairness Metrics}

Here, we describe the fairness metrics used to quantify though not directly optimize trade-offs between utility maximization and fairness outcomes. Specifically, we focus on DP and EO. To define the latter, we make use of the concept of EMC.

\subsection{Demographic Parity}

DP~\cite{caton2024fairness}  ensures that subpopulations (e.g., defined by gender, ethnicity...) are equitably represented in the selected set, even if the groups differ in performance. So, the selection probabilities for different groups will be proportional to their representation in the population. For a policy \(\pol\), it is defined as:
\begin{equation}
| \pol(a_i = 1 | G = g_1) - \pol(a_j = 1 | G = g_2) | \leq \epsilon, \, \forall g_1, g_2 
 \end{equation}
\(G\) is the group membership ( male or female in this work), and \(\epsilon\) is a tolerance parameter. This metric prevents one group from being disproportionately favored.

\subsection{Expected Marginal Contribution}
The EMC is a metric for assessing each candidate's impact on the selection set. It quantifies how much a candidate improves the overall utility when they are selected under a specific policy. This measure is particularly useful where decision makers must evaluate candidates based on uncertain or probabilistic outcomes. Under a fixed policy \( \pol \), the EMC of the candidate \( i \) is defined as:
\begin{align}
\text{EMC}_i(U, \pol) &= \mathbb{E}_{\Pfeat \sim P(\Pfeat)} \mathbb{E}_{\Pact \sim \pol(\Pact \mid \Pfeat)} \big[\left( U(\Pact + i, \Pfeat) - U(\Pact, \Pfeat) \right) \big] \notag\\
&= \sum_{\Pfeat} P(\Pfeat) \sum_{\Pact} \pol(\Pact \mid \Pfeat) \big( U(\Pact + i, \Pfeat) - U(\Pact, \Pfeat) \big) 
\end{align}
in this expression, \( \Pact + i \) represents the selection set \( \Pact \) after adding candidate \( i \). The expectation is taken over both the distribution of feature $P(\Pfeat)$ and the policy selection probabilities $\pol(\Pact \mid \Pfeat)$, to capture the uncertainties in both the candidate pool and the selection process.

This formulation is closely inspired by the EMC definition introduced by \cite{kleine2022meritocracy}. However, an important distinction in our approach is that the EMC is now evaluated under a fixed selection policy while accounting for population uncertainty, and could capture the probabilistic nature of decision-making.  Then, the formulation acknowledges that candidate evaluations may vary due to randomness, selection strategy, or external influences.

\subsection{Equality of Opportunity}
       
Also known as the equal acceptance rate defined in~\cite{verma2018fairness}, EO ensures that individuals with similar EMC qualifications have equal probabilities of selection between groups. Unlike DP that enforces equal selection probabilities  regardless of differences in qualifications or utility contributions (this can result in individuals being selected based on group membership rather than their measured qualifications  i.e., qualification in the observed feature space (e.g., test scores), which may diverge from the underlying construct space institutions ultimately care about (e.g., future potential ) \cite{friedler2021possibility}. Our framework operates on the observed space; bridging this gap is left for future work), EO, however, ensures that individuals with similar qualifications (EMC above a threshold $\tau$) have equal selection probabilities across groups. This avoids penalizing high-performing individuals while still addressing fairness concerns.

\begin{align}
&|\pol (a_i =1 \mid EMC_i \geq \tau , G= g_1) \nonumber\\
 &- \pol (a_j =1 \mid EMC_j \geq \tau , G= g_2)| \leq \epsilon, \forall g_1, g_2
\end{align}

In our utility function, $c$ represents the cost of selecting a set of individuals, with larger values discouraging high admission rates. By setting $\tau = c$, we align the fairness threshold with the admission cost, so that EO is evaluated among the candidates whose expected contribution exceeds the selection cost. 

To incorporate multi-dimensional fairness~\cite{kleinberg2016inherent} penalties into our framework, we merge both fairness metrics into a single overall fairness violation. An approach is to use a weighted sum,  balancing the relative importance of each fairness criterion. In this case, the overall fairness violation \(P_{\text{overall}}\) is defined as
\begin{equation} \label{eq:group}
P_{\text{overall}} = \lambda_{\text{dem}} \, P_{\text{dem}} + \lambda_{\text{eq}} \, P_{\text{eq}},
\end{equation}
 \(P_{\text{dem}}\) is the DP penalty, \(P_{\text{eq}}\) is the EO penalty, and \(\lambda_{\text{dem}}\) and \(\lambda_{\text{eq}}\) are non-negative weights that reflect the importance of each fairness measure. 

An alternative approach is to adopt the maximum operator, which ensures that the most significant fairness violation dominates the overall penalty. In this method, the overall fairness violation is given by
\begin{equation}
P_{\text{overall}} = \max\left( P_{\text{dem}}, \, P_{\text{eq}} \right).
\end{equation}
This approach is particularly useful when it is critical to protect against any severe deviation from fairness, as it directly penalizes the worst-case scenario among the fairness metrics. 

\section{Experiments in the One-shot Setting}
\label{sec:exp}
Experiments are conducted \emph{in simulation}, as the interactive nature of the decision process prevents evaluation on static datasets. We construct a simulator $P^*$ based on a university admissions dataset \cite{kleine2022meritocracy}. This can allow to study realistic admission dynamics while maintaining control over the environment. Using a simulator based on actual data, we can test different policies in a consistent and repeatable way, without needing access to real-time decision-making or sensitive student records.

We evaluate both policy classes introduced in Section~\ref{sec:algorithm}: a \textbf{logistic policy} and a \textbf{neural network policy}. The neural policy is implemented as a feedforward network that maps applicant features to admission probabilities. It consists of two hidden layers with ReLU activations, followed by dropout for regularization, and a final sigmoid output layer producing admission probabilities. In practice, we found that deeper architectures did not provide significant performance gains in our setting, while increasing computational cost and training instability. Two hidden layers were sufficient to capture the relevant nonlinearities in the admission decision problem.

The \textbf{outcome prediction model} ($P(\Pout \mid \Pfeat, \Pact)$) is defined here as a Bayesian multivariate linear regression model~\cite{minka2000bayesian} trained on historical data from previously admitted students and their academic outcomes. To account for uncertainty in future outcomes, we use \emph{posterior predictive sampling} during policy evaluation to sample possible outcomes from the posterior predictive distribution of the model. It lets us evaluate the admission policy using different possible student outcomes, without needing to retrain or change the model.

\subsection{Utility of Logistic vs Neural Policies  }

\begin{figure}[ht]
\centering
\includegraphics[width=1\columnwidth]{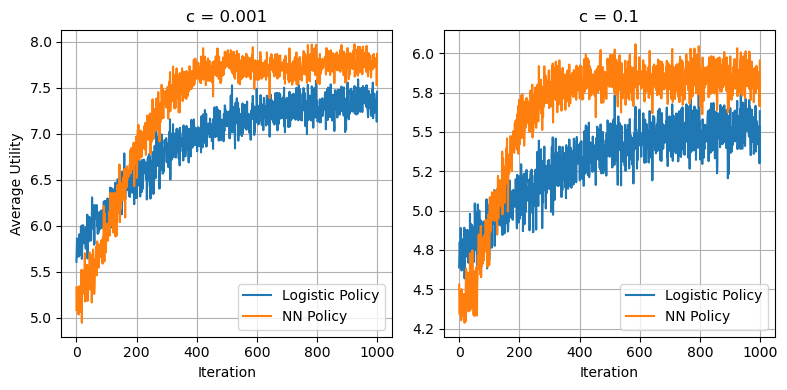}
\caption{Evolution of the expected utility during policy optimization (1000 gradient iterations) for the one-shot setting under varying admission costs $c$.}
\label{fig1}
\end{figure}

To compare the performance of both policies after optimization, We examine how the average utility evolves over optimization iterations for each policy type under two different values of the penalty parameter $c \in \{0.001, 0.1\}$, which controls the trade-off between admitting more applicants and maximizing expected utility. Although decisions are made in a single admission round, the policy itself is learned offline using iterative policy gradient optimization. To ensure robustness, each configuration is repeated in 10 independent trials, and the results are reported in Figure~\ref{fig1}.

When $c = 0.001$, the cost of admission is negligible, and both policies are encouraged to admit more candidates. In this regime, the neural policy converges faster, after an initial exploration phase (approximately 150 iterations), achieves higher utility than the logistic policy. This suggests that the additional flexibility of the neural network allows it to better exploit patterns in the data when the optimization is primarily driven by performance.

When $c = 0.1$, admissions become more costly and require more selective decisions. In this setting, the gap between the two policies becomes more pronounced: the neural policy maintains stable improvements and reaches higher utility levels, while the logistic policy exhibits slower progress and lower final performance. This indicates that, under stricter constraints, the expressiveness of the neural policy becomes particularly beneficial.

Overall, these results show that while both policies improve with training, the neural network consistently provides better utility outcomes, especially in regimes where the decision problem is more constrained. At convergence, the logistic and neural policies admit, respectively, $80.2\%$ and $85.3\%$ of applicants when $c = 0.001$, and $78.3\%$ and $65.7\%$ when $c = 0.1$ . Notably, the logistic policy's admission rate is nearly unchanged between the two cost regimes, whereas the neural policy adapts its selectivity substantially. This result just confirmed what what said early.

\subsection{Effect of Batch Size and Selection Cost $c$} 
\begin{figure}[ht]
  \centering
  \includegraphics[width=1\columnwidth]{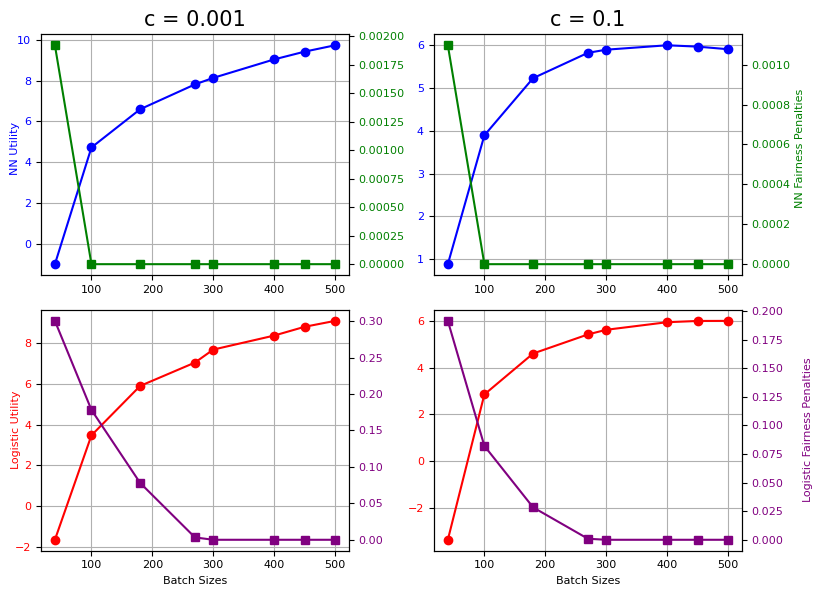} 
  \caption{Expected utility (blue and red curves) and fairness violations (green and purple curves) with respect to batch size for different values of cost $c$.}
  \label{fig2}
\end{figure}

To evaluate the trade-offs between utility and fairness, we systematically vary the batch size and selection cost $c$, and compute utility and fairness penalties for both neural network and logistic policies for the test set. The batch size controls how many applicants are evaluated per iteration. The plots in Figure~\ref{fig2} show how both utility and fairness metrics evolve with these parameters. 

\emph{Impact of $c$ on utility.} When $c = 0.001$, admitting students incurs little cost, and both policies benefit from selecting larger sets of applicants. As the batch size increases, the utility improves steadily, with the neural policy achieving higher values overall. When $c = 0.1$, the cost discourages large admissions, resulting in lower utility levels. In this regime, the neural policy still adapts effectively and maintains an increasing trend, whereas the logistic policy struggles initially and improves more slowly, eventually plateauing at a lower level. This confirms that higher selection costs make the optimization problem more challenging, particularly for less expressive policies. higher costs make optimization more challenging, particularly for smaller batch sizes, where the neural policy exhibits greater robustness. As the batch size increases, the performance gap between the two policies narrows.

\emph{Impact of batch size on fairness penalties.} Fairness penalties drop significantly as batch size increases. At small batch sizes, both policies suffer from noticeable fairness violations, especially the logistic model. The neural network policy shows a sharp improvement: its fairness violation drops rapidly and reaches zero after only 100 student evaluations. In contrast, the logistic policy reduces its fairness violation more gradually, only approaching zero when the batch size exceeds 200. This suggests that expressive models such as neural networks can more efficiently adapt to fairness constraints when provided with sufficient batch-level information, whereas simpler models such as logistic regression require larger batches to achieve similar levels of fairness.

\section{Experiments in the Sequential Setting} \label{mm}

We consider a \emph{multi-stage} (sequential) policy learning setting inspired by real-world decision processes such as university admissions, where policies are revised over time based on observed outcomes. The goal is to iteratively improve decision-making using information collected from previous admission cycles. Each stage corresponds to an admission cycle (year), and the admission policy is updated based on the observed performance of previously admitted students.

The process works as follows: At each stage $t$, a
population $\Pfeat_t = (x_t^1, \ldots, x^n_t)$ of size $n$ is sampled from an underlying (unknown) distribution $P^*(\Pfeat)$. The policy $\pi_{t-1}$ (where $\pi_0$ is an arbitrary policy) is then applied to select a subset of applicants, resulting in an action vector $\Pact_t \in \{0,1\}^n$. Outcomes are generated for all individuals according to the environment, $y_t^i \sim P^*(\Pout \mid x_t^i)$. However, only the outcomes of admitted individuals
(i.e., $a_t^i = 1$) are revealed to the decision-maker. The utility $U(\Pact_t, \Pout_t)$ is then computed based on the selected cohort and their outcomes.

This partial observability reflects the realistic setting in which outcomes for rejected applicants remain unobserved. At the end of each stage, the decision-maker updates both (i) an applicant distribution model $P_t(\Pfeat)$ and (ii) an outcome prediction model $P_t(\Pout \mid \Pfeat, \Pact)$ using all data collected up to stage $t$. These updated models are then used to compute a new policy $\pi_t$ that maximizes the expected utility under the learned distributions, i.e.,
$
\pi_t = \arg\max_{\pi} U(\pi, P_t)
$ which is applied in the next stage.

All experiments are conducted in simulation, as static datasets do not allow evaluation of such interactive and sequential decision processes.

\subsection{Sequential Learning Framework}

We now detail the components of this sequential learning framework, including data generation, model updates, and policy optimization at each stage.

\paragraph{Data generation at each stage.} At the beginning of each stage, a synthetic applicant pool is generated from a simulator $P^*(\Pfeat)$ implemented using a CTGAN, trained on real university admission data. For each applicant, academic outcomes (e.g. GPA in three courses) are generated using a fixed model $P^*(\Pout \mid \Pfeat)$, implemented as a multi-output Bayesian ridge regression model trained on the initial dataset. This outcome generator remains fixed across stages and is used solely to simulate ground-truth outcomes for newly generated applicants. Admission decisions are made based on a learned policy; only the outcomes of admitted students are revealed to the learner.

\paragraph{Model Updates.} At each stage, two models are updated using the accumulated data.
\begin{itemize}

    \item \textbf{Applicant distribution model ($P_t(\Pfeat)$):} 
    We use another CTGAN trained on all observed applicants up to stage $t$ (both admitted and rejected). This model captures the evolving feature distribution of the applicant pool and is used to generate samples for policy optimization.

    \item \textbf{Outcome Prediction Model ($P_t(\Pout \mid \Pfeat, \Pact)$):} We employ a Bayesian multivariate linear regression model \cite{minka2000bayesian} that is continuously updated over stages using the outcomes of admitted students. Unlike a simpler model like basic linear regression~\cite{kleine2022meritocracy}, it maintains a full posterior distribution over regression weights and outcome noise, thus capturing both parameter uncertainty and outcome variability. At each stage, we use \emph{Thompson Sampling} by drawing a single model from this posterior to guide decision-making. This approach naturally balances exploration and exploitation: early stages explore different admission policies via model uncertainty, while later stages exploit accumulated knowledge. The model is retrained after each stage with newly observed data, continuously refining its estimates as more outcomes become available.
   
\end{itemize}

\paragraph{Policy Learning over Admission Stages.} We are using both policy architectures defined in section~\ref{sec:algorithm}.

In the initial stage ($t = 0$), no historical outcome data is available. Therefore, we initialize the process with a simple heuristic policy $\pi_0$, which admits applicants whose predicted average GPA exceeds a fixed threshold. This provides a reasonable starting point before sufficient data are available for learning-based optimization. 

Once the first cohort of students is admitted and their academic outcomes are observed (i.e., from stage $t \geq 1$ onward) , the policy is updated iteratively using a policy gradient approach. At each stage, the decision-maker optimizes the policy parameters to maximize the expected utility under the current models $P_t(\Pfeat)$ and $P_t(\Pout \mid \Pfeat, \Pact)$. The utility includes a penalty term controlled by the parameter $c$, which regulates the trade-off between admission volume and expected outcomes.

The updated policy is then applied to the current applicant pool, and the outcomes are observed for admitted students. These new observations are incorporated into both the applicant distribution model and the outcome prediction model. Over time, this smarter and self-improvement framework allows the university to make progressively well-informed decisions, ensuring that admission strategies remain aligned with university goals while adapting to real-world changes in student quality and reputation.

This sequential procedure reflects realistic admission settings, where decisions influence future data availability and must be continuously refined under uncertainty.

  \subsection {Multi-Stage  Evaluation}

\begin{figure}[ht]
  \centering
  \includegraphics[width=1\columnwidth]{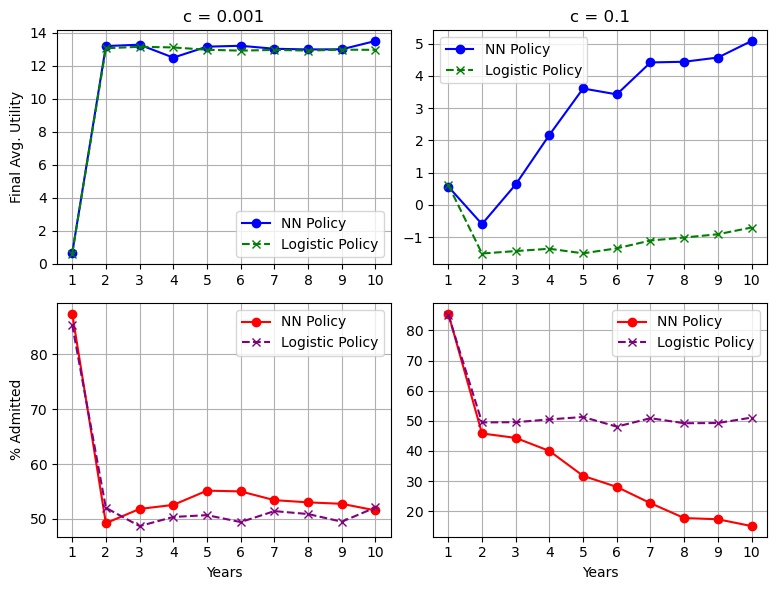} 
  \caption{Final Average Utility and \% of admitted per stage (averaged over 5 trials) under 100 iterations.}
  \label{fig3}
\end{figure}

\paragraph{Utility and Admission Rate.}   To understand how policies evolve over time, we evaluate both the achieved utility and the admission rate at each stage. Tracking these two quantities jointly is important: utility alone does not reveal whether improvements come from better selection or simply from admitting more students, while the admission rate provides insight into how selective the policy becomes under different cost regimes. For each value of the penalty parameter $c$, the results are averaged over 10 trials to reduce variability.

The upper row of Figure~\ref{fig3} shows the evolution of utility between admission cycles, while the lower row reports the corresponding admission rates. When $c = 0.001$, admission is weakly penalized, and both policies quickly converge to high and stable utility values (above 12), with admission rates around 50--55\%. This indicates that in low-cost settings, even simple models are sufficient to maintain strong performance.

In contrast, when $c = 0.1$, the setting becomes more constrained and the differences between policies become more pronounced. The neural network policy continues to improve over stages, while the logistic policy struggles, often remaining close to zero or even having negative utility. At the same time, neural policy becomes increasingly selective, with admission rates gradually dropping below 20\% by later stages. This behavior suggests that the neural policy is able to adapt its selectivity to the cost constraint, focusing on high-value candidates, whereas the logistic policy lacks the flexibility to adjust effectively. In a university admissions context, this translates into a more strategic and data-driven selection process when using expressive models.

\begin{figure}[ht]
  \centering
  \includegraphics[width=0.45\columnwidth]{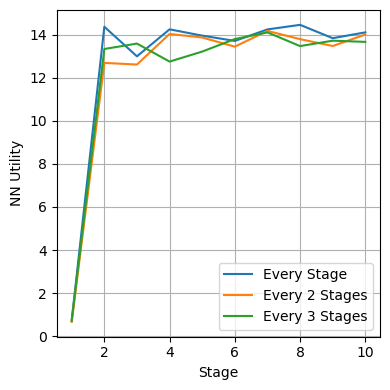} 
  \includegraphics[width=0.45\columnwidth]{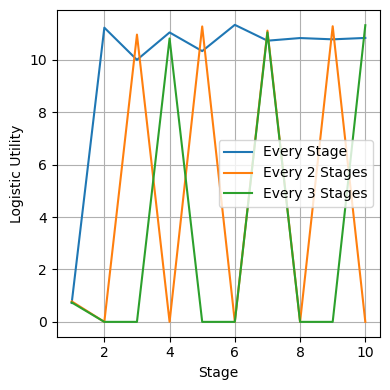} 
  \caption{Average utility across stages under different policy update frequencies under 50 iterations and c=0.001.}
  \label{fig4}
\end{figure}

  \paragraph{Delay policy updated.} We further investigate how frequently the admission policy needs to be updated. In practice, universities may not revise their admission criteria every year due to administrative or operational constraints. To reflect this, we evaluate the impact of delayed updates by comparing policies updated every stage with policies updated every 2 or 3 years (Figure~\ref{fig4}). This experiment helps assess whether frequent retraining is necessary or if policies remain effective when updated less often, balancing between adaptability and practicality.

The results show a clear difference between the two policy classes. The neural network policy (left) remains relatively stable even when updates are less frequent, maintaining strong utility across stages. This suggests that more expressive models can generalize better over time and do not require constant retraining. In contrast, the logistic policy (right) is much more sensitive to delayed updates: skipping updates leads to noticeable drops in performance at several stages. This indicates that simpler models rely more heavily on frequent adjustments to remain aligned with the evolving data distribution. However, we observe that when updates are delayed for too long, the neural policy can start to degrade after several stages (around stage 5), suggesting that although neural policies are robust to moderate delays, excessively infrequent updates lead to unstable performance and poor generalization over time. Additional results illustrating this behavior are provided in the Appendix \ref{nn}.

From a practical perspective, these results show an important trade-off. Although frequent updates improve performance, they also require more resources. Neural policies offer a useful advantage here, as they allow decision-makers to update less often without significantly degrading performance, making them more suitable for real-world deployment in settings like university admissions.

  \paragraph{Compare against static baselines.} 
To assess the value of adaptive decision-making, we compare our sequential policies with several static baselines that remain fixed at all stages. This comparison is important in practice, as many institutions rely on fixed admission rules due to their simplicity and ease of implementation. However, such approaches do not adapt to newly observed data, which may limit their long-term effectiveness.

We consider three static baselines: (i) a logistic policy trained once and kept fixed, (ii) a neural network policy trained once and reused without updates, and (iii) a threshold-based policy that admits students whose predicted GPA exceeds a predefined cutoff. These baselines allow us to isolate the benefit of updating the policy over time.

    \begin{figure}[ht]
  \centering
  \includegraphics[width=1\columnwidth]{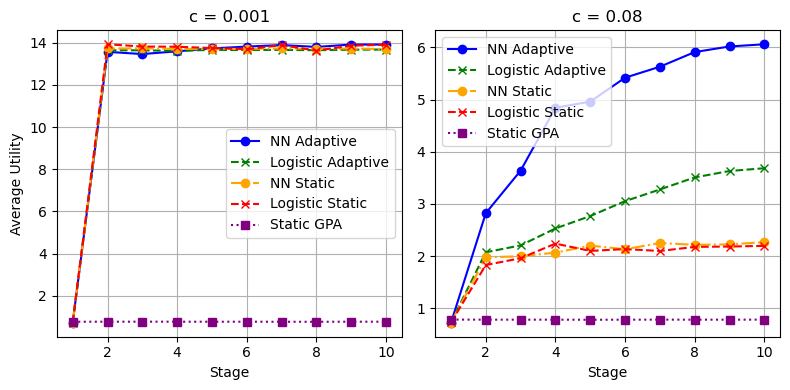}  
  \caption{Comparison of adaptive and static admission policies across stages for two cost settings. }
  \label{fig5}
\end{figure}

A commonly used benchmark, such as top-$k$ selection is not included here, as it assumes access to true outcomes for all applicants. This assumption is unrealistic in admission settings, where outcomes are only observed for admitted students. In contrast, our framework explicitly accounts for this uncertainty, making the comparison more aligned with real-world decision-making. All results are averaged over 10 trials.

The results in Figure~\ref{fig5} highlight the importance of adaptivity in sequential decision-making. When the admission cost is low ($c = 0.001$), all learning-based policies quickly reach a similar level of high utility (around 14). This suggests that in settings where admitting students is not strongly penalized, even fixed policies are sufficient to maintain good performance. The only clear outlier is the GPA threshold baseline, which remains significantly lower, indicating that simple heuristics do not capture the complexity of the decision problem.

The difference becomes much more pronounced when the cost increases ($c = 0.08$). In this more constrained setting, adaptive policies clearly outperform their static counterparts. The neural adaptive policy shows the strongest improvement over stages, steadily increasing its utility, and reaching the highest overall performance. The logistic adaptive policy also improves over time, but at a slower rate and with a lower final utility. In contrast, both static policies (logistic and neural) plateau early and fail to adapt to dynamic data, resulting in consistently lower performance. The GPA-based policy again remains far below all other methods. We use $c = 0.08$ rather than $0.1$ in this comparison, as $c = 0.1$ led to near-zero admission rates for several baselines, obscuring the comparison of adaptive vs. static policies.

In a university admissions context, this means that relying on fixed admission rules can lead to suboptimal outcomes, especially when selectivity matters. Adaptive approaches, particularly those based on more expressive models, such as neural networks, are better able to refine their strategy over time and make more effective  decisions.

\begin{figure}[ht]
  \centering
  \includegraphics[width=1\columnwidth]{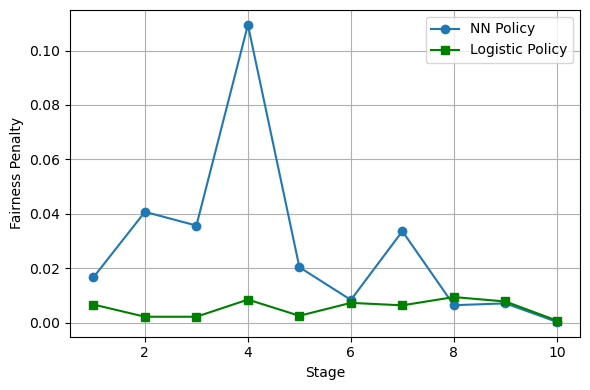}  
  \caption{fairness violation  across stages for neural network  and logistic policies under 50 iterations and $c=0.001$. }
  \label{fig6}
\end{figure}

\paragraph{fairness violation.}  In addition to utility, we evaluate how fairness evolves across stages of the decision process.  We measure fairness using standard group-based criteria, including DP and EO ( Eq. \ref{eq:group} with $\lambda_{\text{dem}} = \lambda_{\text{eq}}=1$), computed at each stage based on the current policy and applicant pool (Figure~\ref{fig6}).

Both policies maintain relatively low general fairness penalties, which is reassuring given that fairness is not explicitly optimized. The neural policy exhibits more variability in early stages, likely due to exploration and model uncertainty, but stabilizes quickly and reaches a near-zero penalty in later stages. In contrast, the logistic policy remains more stable throughout, though with smaller fluctuations that persist across stages. Overall, improving performance does not systematically come at the expense of fairness, as both policies maintain low fairness penalties across stages, with the neural policy progressively reducing violations as it adapts to new data.

\section{Discussion and Future Works}

This work studies adaptive decision-making under uncertainty in the context of university admissions, highlighting the benefits of learning policies that evolve over time using a policy gradient framework. Our results show that optimizing admission policies with respect to expected utility leads to strong performance while maintaining reasonable fairness outcomes even when fairness is not explicitly enforced. In particular, neural network policies demonstrate greater flexibility and robustness compared to simpler models such as logistic regression, especially in dynamic and constrained settings. We showed that the cost parameter $c$ (penalization of admitting too many students while having small resources) can play a key role in the admission process. The experiments also illustrate that, under higher admission costs, the learned policy naturally becomes more selective. While this improves the defined utility, it also highlights that admission volume is itself an institutional design choice rather than a purely technical consequence. Different universities may prefer different trade-offs between cohort size and expected academic outcomes, depending on their educational mission and available resources.

Specifically, through multi-stage experiments, we looked at cases where policy updates are delayed, and found that even with less frequent updates, neural network still performs well  but not logistic policy. A particularly noteworthy finding is the consistently poor performance of the threshold-based admission policy. Although threshold rules based on predicted GPA are widely used because of their simplicity and transparency, our experiments show that such static rules fail to adapt to changing applicant populations and therefore produce substantially lower utility than adaptive policies. Among all approaches tested, the neural network-based policy consistently outperformed others, including the logistic one, showing a stronger ability to adapt and improve over time. 

We note that fairness in this work is treated primarily as an evaluation criterion rather than a design constraint: our policies are optimized solely for expected utility, and DP/EO are measured post-hoc rather than enforced during training, which is a limitation. Extending the framework in Section \ref{sec:algorithm} with an explicit fairness-constrained is a natural and important next step. Additionally, expanding this work to online learning as sequential decision-making where we update the policy incrementally as each applicant is evaluated, rather than waiting for a full annual cycle.  An interesting direction for future work is to relate the proposed framework to causal policy learning methods, which explicitly model counterfactual outcomes and treatment effects. While our work focuses on sequential policy optimization under partial feedback using probabilistic predictive models, integrating causal estimators could provide additional robustness in settings where confounding plays an important role.

\bibliography{sequential}

\appendix

\section{Dataset Description}

All experiments are conducted using data generated from a simulator $P^*$ trained on data from the Norwegian university admissions database~\cite{kleine2022meritocracy}. The use of a simulator is \emph{necessary} due to the interactive nature of the problem, as evaluation cannot be performed on static datasets where outcomes are only observed for admitted students. This approach allows us to model realistic admission dynamics while maintaining full control over the experimental environment. It also allows for consistent and repeatable evaluation of different policies without requiring access to sensitive or real-time student data.

The original dataset is not publicly available, as it was obtained under a data-sharing agreement for research purposes. It includes academic features such as high school GPA, science scores, language scores, and additional admission-related attributes. Admission decisions are binary (0 = rejected, 1 = admitted), with an approximate acceptance rate of 70\%.

To enhance realism, we train a Conditional Tabular Generative Adversarial Network (CTGAN)~\cite{xu2019modeling} for 100 epochs and generate 1,000 synthetic applicants. This ensures that the generated data remains statistically consistent with the original distribution. From these applicants, we identify the admitted students and simulate three course-specific GPA outcomes based on their academic profiles. These GPAs are clipped to lie within the range $[0,4]$.

We preprocess the data by applying one-hot encoding to categorical variables and normalizing numerical features using Min-Max scaling. The generated course GPAs are also normalized to ensure consistency across models. Finally, the dataset is split into 70\% training and 30\% testing subsets for robust evaluation.

\paragraph{Experimental Setup.} All experiments were conducted on a MacBook Pro equipped with an Apple M2 chip and 8\,GB of RAM, running macOS Ventura. The implementation relies on Python, using PyTorch (v2.0), NumPy (v1.24), scikit-learn (v1.2), and the CTGAN model from the SDV library (v0.7.0). External GPU acceleration was not used. The experimental setup remains lightweight and fully reproducible on standard personal hardware.

 Real-world admission cycles at large institutions can involve tens of thousands of applicants. The main computational bottlenecks in our framework are (i) CTGAN retraining at each stage, and (ii) the Monte Carlo estimation of the policy gradient, whose cost scales with the number of population samples $n_x$, action samples $n_a$, and population size $N$. We expect the policy optimization itself to scale reasonably (batched gradient computation is standard in PyTorch), but CTGAN retraining at every admission cycle could become a bottleneck at larger $N$ or with more frequent stage updates; Investigating scalability on institutional-scale datasets constitutes an interesting direction for future work.
 
\section{Algorithms} \label{app:algo}

In this section, we describe the practical implementation of the proposed policy optimization framework. We consider two settings. First, the \emph{one-shot} setting where the policy is learned on a fixed population distribution. Second, a \emph{multi-stage} setting where the population evolves over time and both the policy and the outcome model are updated iteratively. In both cases, we rely on Monte Carlo sampling to approximate the expected utility and optimize the policy using gradient-based methods.

\subsection{One-shot setting}

In the one-shot setting, the population distribution $P(\Pfeat)$ is assumed to be fixed and known through samples. The goal is to learn a policy that maximizes expected utility on this static population. The Algorithm~\ref{alg:sg_fixed} starts from an initial parameter vector ($\theta_0$), the algorithm iteratively updates the policy by sampling batches of applicants from the population distribution and generating admission decisions according to the current policy. For each sampled batch, multiple action vectors are drawn to approximate the expectation over ( $\pol_\theta(\Pact \mid \Pfeat)$ ). The corresponding utilities are then used to compute a Monte Carlo estimate of the policy gradient, following the formulation in equation \ref{eq:grad}. Gradient estimates are averaged across samples to reduce variance, and policy parameters are updated using gradient ascent (or equivalently, gradient descent on the loss). The process is repeated until convergence.

\begin{algorithm}[H]
\caption{Policy Gradient for single-shot setting}
\label{alg:sg_fixed}
\begin{algorithmic}[1]
\State \textbf{Input:}  population distribution $P(\Pfeat)$, model $P(\Pout|\Pfeat, \Pact)$, utility function $u$, sample size $M$, batch size $B$,learning rate $\eta$, convergence threshold $\delta$
\State \textbf{Initialize:} Policy parameters $\theta^0$, 

\While{$\|\theta^{t} - \theta^{t-1}\| > \delta$}
    \State Sample a batch of applicants $\Pfeat_B \sim P(\Pfeat)$ 
    \State Initialize gradient accumulator $G \leftarrow 0$
    
    \For{$k = 1$ \textbf{to} $M$} \Comment{Monte Carlo Sampling Loop}
        \State Evaluate $\nabla_{\theta_1} U(\pol_{\theta_1}, \Pfeat)$ or $\nabla_{\theta_2} \mathcal{L}(\theta_2)$ from $u, \Pfeat$ and $P$
        \State Accumulate gradient: $G \leftarrow G + \frac{1}{M} ( \nabla_{\theta_1} U(\pol_{\theta_1}, \Pfeat)$ or $G \leftarrow G + \frac{1}{M} ( \nabla_{\theta_2} \mathcal{L}(\theta_2))$
    \EndFor
    
    \State Update Policy Parameters: 
    $\theta^{t+1}_1 \leftarrow \theta^t_1 + \eta G$ or $\theta^{t+1}_2 \leftarrow \theta^t_2 - \eta G$
    
\EndWhile
\State \Return $\pi_{\theta^*}$
\end{algorithmic}
\end{algorithm}

\subsection{Multi-stage setting}

In the multi-stage setting, the decision process unfolds over multiple stages, where the applicant population evolves over time. At each stage, a new cohort of applicants is observed, decisions are made using current policy, and the results are only partially revealed (for selected individuals). The outcome model and policy are then updated based on the accumulated data. 

To capture the evolving population, we use a generative model to simulate new applicants at each stage. The policy is optimized iteratively using policy gradient updates, while the outcome model is refined using newly observed data. This setting reflects realistic scenarios, such as university admissions, where decisions influence future data and must adapt over time.

\begin{algorithm}[H]
\caption{Multi-Stage Policy Gradient with Dynamic Updates}
\label{alg:sg_1}
\begin{algorithmic}[1]

\State \textbf{Input:}  number of stages $K$, base learning rate $\eta$, decay factor $\gamma$,sample size $M$, batch size $B$
\State \textbf{Initialize:} Policy $\theta^0$, simulator ($P^*(\Pfeat),P^*(\Pout \mid \Pfeat)) $, distribution model $P_t(\Pfeat)$, Bayesian outcome predictor $P_t(\Pout \mid \Pfeat, \Pact)$, Applicant History $\mathcal{H} \leftarrow \emptyset$, Training Data $\mathcal{T} \leftarrow \emptyset$

\For{$t = 0$ \textbf{to} $K-1$}
    \State $\eta_t \leftarrow \eta / (1 + \gamma \cdot t)$ \Comment{Apply learning rate decay}
    
    \State \textbf{Step 1: Environment Simulation}
    \State Sample synthetic applicant pool $\Pfeat_t \sim P^*(\Pfeat)$
    \State Predict latent outcomes $\Pout_t \sim P^*(\Pout \mid \Pfeat)$

    \If{$t == 0$}
        \State $\Pact_t \leftarrow$ Apply threshold-based heuristic $\pi_0(\Pfeat_t)$
    \Else
        \State \textbf{Step 2: Policy Optimization }
        \State Sample posterior parameters $\theta_{TS}$ from $P_t(\Pout \mid \Pfeat, \Pact)$ \Comment{Thompson Sampling}
        \State Run Algorithm \ref{alg:sg_fixed} using $\theta_{TS}$
        \State $\Pact_t \leftarrow$ Apply learned policy $\pi_{t-1}(\Pfeat_t)$
    \EndIf
    
    \State \textbf{Step 3:  Data Collection}
    \State Observe outcomes $\Pout_{obs}$ for admitted individuals
    \State $\mathcal{D}_{t+1} \leftarrow \mathcal{D}_t \cup \{\Pfeat_{admit}, \Pout_{obs}\}$
    \State $\mathcal{H}_{app} \leftarrow \mathcal{H}_{app} \cup \Pfeat_{t}$ \Comment{Update applicant history}
    
    \State \textbf{Step 4: Model Refinement}
    \State Update applicant model $P_t(\Pfeat)$ on $\mathcal{H}_{app}$
    \State Update Bayesian predictor $P_t(\Pout \mid \Pfeat, \Pact)$ with new data $\mathcal{D}_{k+1}$
\EndFor

\State \Return Final optimized policy $\pi_{\theta^*}$

\end{algorithmic}
\end{algorithm}

\section{Simple Baseline comparison for the one-shot setting}

\begin{figure}[ht]
  \centering
  \includegraphics[width=0.8\columnwidth]{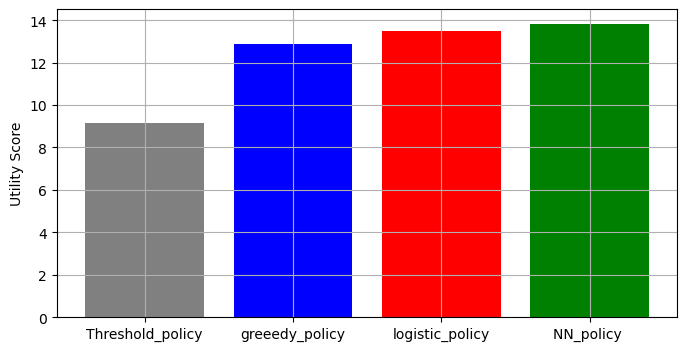} 
  \caption{Comparison of utility scores for different selection strategies: threshold-based selection, greedy selection, and Logistic and neural network-based selection.}
  \label{fig7}
\end{figure}

  To evaluate the effectiveness of our policy utility in the one-shot setting, we compare it against two baseline selection methods: Threshold-based Selection and Greedy Selection. These baselines provide useful benchmarks to assess how well our policy approaches optimize admissions decisions in figure \ref{fig7}.

 \paragraph{The Threshold-based Selection.} This approach admits candidates whose predicted GPA or performance in three key courses exceeds a predefined threshold. It ensures that only applicants meeting a minimum academic standard are selected, offering a straightforward but rigid method that does not consider the overall distribution of applicants' abilities. Although it maintains academic quality, it can exclude candidates who could perform well despite falling just below the threshold. 

 \paragraph{The Greedy Approach (Based on Total Outcome).} It prioritizes the applicants with the highest predicted academic performance by selecting those with the highest sum  of grades across all courses. Unlike the threshold method, which applies a fixed cutoff point, the greedy approach ranks candidates and selects the top performers, assuming that maximizing total academic achievement will yield the best outcomes. However, this method may favor candidates with strong overall scores while neglecting other factors that contribute to long-term success, such as diversity in skills or adaptability to different learning environments. In both methods, the number of selected applicants remains the same. 

The Neural network approach achieves the highest utility, showing that it optimizes student selection more effectively than all other methods. The greedy approach, which selects individuals based on total predicted outcomes, performs better than the threshold method, suggesting that a naive optimization of total scores could yield the best long-term outcomes. The logistic policy can also be seen as a baseline because it has been used and implemented in \cite{kleine2022meritocracy} for the single-shot setting.

\section{Effect of Highly Delayed Policy Updates for neural network} \label{nn}

\begin{figure}[ht]
  \centering
  \includegraphics[width=0.7\columnwidth]{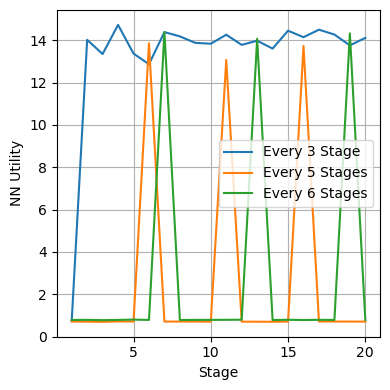} 
  \caption{Average utility across stages under different policy update frequencies under 50 iterations and c=0.001.}
  \label{fig8}
\end{figure}

We further examine the behavior of the neural policy under more extreme update delays, where the policy is updated only every 3, 5, or 6 stages. Figure \ref{fig8} shows that while the policy can still achieve high utility when updated every 3 years, performance quickly deteriorates when update periods are very late. In particular, we observe that when updates are delayed for too long, the neural policy can start to degrade after several stages (from stage 5), suggesting that although neural policies are robust to moderate delays, excessively infrequent updates lead to unstable performance and poor generalization over time.

\end{document}